\documentclass[journal]{IEEEtran}
\usepackage{amsmath}
\usepackage{amssymb}
\usepackage{algorithm}
\usepackage{algorithmic}
\usepackage{graphicx}
\usepackage{tikz}
\usetikzlibrary{arrows.meta,positioning,fit}
\usepackage{orcidlink}
\hypersetup{
    colorlinks=true,
    linkcolor=blue,
    citecolor=blue,
    urlcolor=blue
}

\ifCLASSINFOpdf
\else
\fi
\begin{document}
\raggedbottom
%
\title{Feedback-Driven Prompt Refinement for Educational AI: Affective Trajectory Estimation and Bloom’s Taxonomy Classification}
%
%
%

\author{
\IEEEauthorblockN{Josh Dean\orcidlink{0009-0009-9551-2263}, \IEEEmembership{Member, IEEE}; Karan Patel; Gaurangi Raul; Yu-Zheng Lin\orcidlink{0000-0001-7975-3694}, Member, IEEE; Jesus Pacheco\orcidlink{0000-0002-8636-5902}; Sicong Shao\orcidlink{0000-0003-1205-3890}, Member IEEE; Shalaka Satam\orcidlink{0000-0003-0848-4727}; Soheil Salehi\orcidlink{0000-0001-5998-8795}, Senior Member, IEEE; Pratik Satam\orcidlink{0000-0003-3139-8333}
}
}

\markboth{IEEE Transactions on Education}{Refining Educational AI: Prompt Optimization Across Affective and Cognitive Tasks }
%



\maketitle

\begin{abstract}\begingroup\bfseries
\textit{Contribution:} This study has shown that a shared prompt-refinement procedure, guided by distinct affective and cognitive feedback signals, can improve held-out reference agreement and multi-label classification while reducing affective-generation variability without executor parameter updates.
\par\noindent\hspace*{1em}\textit{Background:} Industry 4.0 has increased demand for personalized technical education across interconnected computing and engineering domains, while prompt sensitivity and generation variability have complicated dependable interpretation of instructional dialogue and learning outcomes.
\par\noindent\hspace*{1em}\textit{Research Questions:} The study has examined whether task-specific feedback can improve affective-reference agreement, generation consistency, and cognitive classification without updating executor parameters.
\par\noindent\hspace*{1em}\textit{Methodology:} A secondary language model has revised executor prompts using executor-output discrepancies against model-generated affective references for four dialogues and annotated Bloom labels for 70 learning outcomes; separate sets of 100 EduTalk conversations and 1,000 outcomes have supported held-out comparisons. Small refinement sets, unvalidated affective references, and absent learner-outcome measures have constrained interpretation.
\par\noindent\hspace*{1em}\textit{Findings:} Affective mean absolute error has decreased from 0.742 to 0.702 and mean generation standard deviation from 0.307 to 0.238, while Bloom micro-F1 has increased from 0.687 to 0.893 and exact-match subset accuracy from 0.418 to 0.636. These findings have supported prompt refinement as a candidate approach for developing adaptive technical instruction, while its effects on workforce learning remain to be established.
\par\endgroup
\end{abstract}


%
\IEEEpeerreviewmaketitle

\section{Introduction}
%
%
%
%

\IEEEPARstart{T}{he} Fourth Industrial Revolution integrates cloud infrastructure, Internet of Things (IoT) networks, and smart manufacturing into cyber-physical environments \cite{lee2010cps} that require workers to connect knowledge across networking, systems programming, embedded systems, and other interdependent domains. Industry 4.0 systems become more tightly coupled. Workforce education must accommodate varied prior preparation and learning pace while helping learners apply technical concepts across authentic contexts \cite{satam2023cpstr}. Adaptive educational systems therefore need to coordinate instructional-content access, course-specific knowledge, learner-state representations, and language-model-mediated response generation. Prior evaluations have examined complementary functions that inform such architectures. Slide-based instructional assistance supported on-demand explanation, course-document retrieval grounded tutoring responses in domain materials, and dialogue analyses represented affective and cognitive learner states \cite{raul2025ragprism,lin2026llmmcaffect,patel2025ocr}. These findings motivate the integrated architecture presented here, while the current evaluation tests prompt refinement separately on held-out tasks. A focused question remained whether systematic prompt refinement improved a fixed executor's behavior across distinct pedagogical objectives, including continuous affective estimation and structured cognitive classification. In Industry 4.0 workforce education, the need to connect technical knowledge with equipment operation, fault diagnosis, and system-level reasoning has motivated instructional approaches sensitive to both the cognitive demands of learning objectives and the difficulties learners express in dialogue. Although large language models have expanded opportunities to interpret these complementary dimensions, sensitivity to instructional wording and variability across generations have raised questions about the dependability of the resulting analyses. These concerns have motivated the present investigation of prompt refinement as a means of strengthening dialogue-based affective estimation and cognitive classification within the broader development of adaptive technical education.

This study has evaluated a shared prompt-refinement procedure for utterance-level affective trajectory alignment and multi-label Bloom's Taxonomy classification, using task-specific feedback to revise executor instructions without updating model parameters. Separate refinement and held-out sets have supported comparisons between initial and selected prompts, with model-generated trajectories supplying affective references and annotated learning outcomes supplying cognitive targets. The findings have indicated improved reference agreement, classification performance, and affective-generation consistency; the study has not assessed student learning, retention, or achievement. Evaluating these complementary tasks within one framework has characterized the potential and evidentiary limits of prompt refinement for adaptive technical education.

\section{Related Work}
Research on adaptive educational systems has treated personalization as coordination among instructional content, learner-state estimates, and tutoring behavior in response to individual needs and contexts \cite{duplooy2024adaptive}. Large language models (LLMs) have broadened opportunities for on-demand explanation and feedback, while raising questions about educational use \cite{kasneci2023chatgpt}, learner-state reliability, and control of model behavior without retraining.

Earlier work has combined digital-twin environments, retrieval-assisted tutoring, and zero-shot dialogue sentiment analysis \cite{lin2025personalized}, while subsequent developments have specified retrieval-grounded tutoring through RAG-PRISM \cite{raul2025ragprism} and an emotion-aware PRISM architecture integrating dialogue, digital-twin observations, sentiment analysis, and knowledge-augmented guidance \cite{lin2026prism}. These developments have motivated coordinated consideration of affective and cognitive information; the present study has evaluated prompt refinement on separately defined tasks.

Supporting studies have examined slide-based explanation, course-document retrieval, and equipment reconstruction as complementary foundations for adaptive technical learning. An instructional assistant has simplified slide content in a live course, with perceived-usefulness and single-slide processing comparisons supporting feasibility rather than learning gains or course-wide efficiency \cite{patel2025ocr}. RAG-PRISM has evaluated retrieval and answer quality on generated and perturbed queries \cite{raul2025ragprism}; Table~\ref{tab:architecture_evidence} retains the reconciled retrieval count and reported answer scores, whose verification has been limited by inconsistent published tabulations. Alhamadah et al. have reconstructed industrial equipment from consumer-device imagery, LiDAR scans, and 3D processing, comparing dimensions against manual measurements \cite{alhamadah2024photogrammetry}. These findings have informed the conceptual architecture in Fig.~\ref{fig:adaptive_arch}, while their interventions and outcomes have remained separate from the present prompt-refinement experiments.

\begin{table*}[!t]
\caption{Educational Evidence for Adaptive-Instruction Functions and the Present Evaluation}
\label{tab:architecture_evidence}
\centering
\footnotesize
\begin{tabular}{p{0.16\textwidth}p{0.47\textwidth}p{0.29\textwidth}}
\hline
\textbf{Architectural function} & \textbf{Evidence informing the architecture or evaluated here} & \textbf{Scope of that evidence} \\
\hline
Instructional-content access & In a live technical course, 42 student responses yielded a mean usefulness rating of 7.83/10. A one-slide comparison reported 392 versus 1,105 input tokens for text-based versus image-based processing, with visual processing stronger on visually dense slides \cite{patel2025ocr}. & Perceived usefulness and a single-slide comparison supported feasibility. Neither measured learning gains nor system-wide cost. \\
Course-knowledge grounding & An immersive tutoring prototype combined learner context with course-document retrieval. It retrieved supporting material for 10 of 11 queries (90.9\%). Table IV in RAG-PRISM reported GPT-4 relevancy of 0.933 and faithfulness of 1.000 \cite{raul2025ragprism}. & Small-query, model-judged summaries. Conflicting published tabulations limit verification. These measures do not establish learner outcomes or the efficacy of prompt refinement. \\
Learner-state representation & The present study used the EduTalk dataset of approximately 1,280 teacher--student dialogues and evaluated prompt-refined sentiment trajectory estimates on held-out conversations. & Dialogue-based affect estimation provided a candidate state signal. It did not establish that sentiment estimates are human-validated or that adaptation improves learning. \\
\hline
\end{tabular}
\end{table*}

\subsection{Affective and Cognitive Learner-State Modeling} Adaptive instruction depends on representations of both cognitive objectives and affective change over an interaction. Bloom's Taxonomy provides a framework for describing the cognitive complexity of learning objectives \cite{krathwohl2002revision}. Affective modeling, in turn, has represented utterance-level states as trajectories whose variation can be analyzed across repeated model inferences. The present study brought these complementary tasks into a common prompt-refinement evaluation. LLM-MC-Affect modeled utterance-level affect as a latent distribution, used repeated stochastic inferences to estimate central tendency and dispersion, and derived dyadic interaction indicators from sequential cross-correlation, estimated lag, and trajectory slopes \cite{lin2026llmmcaffect}. Its evaluation compared six LLMs, used 20 Monte Carlo samples per dialogue turn, and examined temperature sensitivity and instructional interaction patterns. The analyses showed that mean trajectories remained comparatively stable as decoding temperature changed, while estimated dispersion increased, and that model families differed in their ability to represent affective turning points. This work therefore showed how affective ambiguity and interpersonal coupling can be analyzed beyond deterministic sentiment labels. At the same time, its controlled case studies used a simulated instructional-dialogue dataset. Its variance measure may conflate linguistic ambiguity with model and decoding effects, and its cross-correlation indicators did not establish causal influence or alignment with human-rated affect. Those qualifications were especially relevant when model-derived affective estimates were reused as feedback signals for adaptive systems.

Related research on reputation analysis has distinguished sentiment expressed in text from the text's implications for a target entity, demonstrating that negative sentiment can coexist with positive reputation polarity \cite{rahman2022bert}. This distinction provides a useful conceptual caution for educational affect modeling, where agreement with a polarity reference has not independently established the validity of an inferred learner state.

\subsection{Behavioral Representation Across Domains}
Outside education, behavioral monitoring has used reference models and complementary observations to interpret system activity. A multi-level Bluetooth intrusion-detection system has represented protocol activity through observation flows and n-grams to distinguish normal from abnormal behavior \cite{satam2020multilevel}. At the same time, multi-layer anomaly detection has combined system, process, and network observations in a shared representation \cite{shao2021multilayer}. These studies provide methodological analogies for reference-based evaluation and contextual representation. However, they do not establish educational validity or the effectiveness of prompt refinement.

\subsection{Automated Prompt Optimization and the Present Study}
Research on prompt sensitivity has shown that meaning-preserving changes in prompt formatting can affect language-model performance in few-shot evaluations \cite{sclar2023quantifying}, raising methodological questions about the dependability of results from a single prompt presentation. For educational classifications and dialogue-based estimates, this evidence has motivated attention to prompt presentation alongside task-specific instruction design. However, the present study has compared initial and selected prompts without isolating sensitivity to formatting or other presentation features.

Prompt engineering often relies on manual trial and error. In contrast, automated approaches treat natural-language instructions as candidates that can be generated, tested, and refined against task-specific feedback. Automatic Prompt Engineer and Optimization by PROmpting (OPRO) have used language models to generate and evaluate candidate instructions without updating executor parameters \cite{zhou2022large,yang2023large}. Self-Refine has instead used iterative self-feedback to revise generated outputs, providing a related feedback mechanism rather than an equivalent prompt-optimization procedure \cite{madaan2023self}. These approaches have informed feedback-guided refinement, although their objectives and evaluation settings have not necessarily reflected the pedagogical criteria and learner-context constraints of adaptive education. Education-oriented post-training has used authentic learning data to shape model behavior \cite{perczel2025teachlm}. Course-specific knowledge provision supplies external evidence at inference time. At the same time, prompt-level refinement remained complementary because it targeted how an executor interpreted task instructions and feedback without changing model parameters or requiring the prompt itself to encode all domain knowledge. Related research on generative digital twins has adapted LLM-based forecasts by updating structured observation windows and aggregating repeated predictions without task-specific parameter training \cite{lin2026dddgendt}. DDD-GenDT identifies adaptive prompt optimization as a future direction. This study examined feedback-driven instruction revision for educational tasks, with industrial forecasting remaining a separate application requiring evaluation.

Prior research on immersive technical learning, course-grounded tutoring, and dialogue-based learner modeling has provided complementary foundations for adaptive Industry 4.0 workforce education, linking access to technical knowledge with contextual interpretation of instructional interactions \cite{lin2026prism,raul2025ragprism,lin2026llmmcaffect}. These developments have motivated investigation of the model behavior underlying two educational functions: estimating affective trajectories during dialogue and characterizing the cognitive demands of learning outcomes across interconnected technical domains. In this context, the present study examined a shared prompt-refinement procedure using distinct feedback signals: model-generated references for affective estimation and annotated labels for multi-label Bloom’s Taxonomy classification. Held-out evaluation assessed reference agreement, classification performance, and generation variability, providing task-level evidence relevant to developing adaptive technical instruction while leaving the relationship between these improvements and workforce learning outcomes open to subsequent investigation.

\section{Methodology}
This study has examined whether a shared feedback-driven prompt-refinement procedure improves affective trajectory estimation and cognitive classification while executor parameters remain fixed. Initial and selected prompts have been compared on held-out measures of reference agreement, classification performance, and generation variability. Prior educational studies have informed the broader architecture, but their datasets, interventions, and outcomes have not been pooled with the present experiments. The design has evaluated model behavior rather than instructional efficacy or learner achievement.

\subsection{Integrated Adaptive-Learning Architecture}
\begin{figure*}[!t]
\centering
\begin{tikzpicture}[font=\footnotesize, stage/.style={draw,rounded corners,align=center,text width=2.65cm,minimum height=1.15cm,inner sep=5pt}, feedback/.style={stage,draw,dashed,fill=orange!10}, flow/.style={-{Latex[length=2mm]},thick}]
\node[stage,fill=gray!10] (input) at (0,0) {\textbf{Learning activity}\\Course material, learner turns, and interaction traces};
\node[stage,fill=blue!8] (prep) at (3.50,0) {\textbf{Content and knowledge context}\\Text or visual input; relevant course excerpts};
\node[stage,fill=green!8] (state) at (7.00,0) {\textbf{Learner-state representation}\\Affective trajectory and cognitive objective profile};
\node[stage,fill=blue!8] (executor) at (10.50,0) {\textbf{Executor model}\\Current prompt, course context, and learner-state cues};
\node[stage,fill=gray!10] (response) at (14.00,0) {\textbf{Instructional response}\\Context-sensitive explanation or feedback};
\draw[flow] (input) -- (prep); \draw[flow] (prep) -- (state); \draw[flow] (state) -- (executor); \draw[flow] (executor) -- (response);
\node[feedback] (evaluation) at (9.20,-2.15) {\textbf{Task evaluation}\\Reference trajectories or annotated labels; task-specific metrics};
\node[feedback] (refiner) at (12.85,-2.15) {\textbf{Prompt refiner}\\Uses observed discrepancies to propose an updated prompt};
\draw[flow] (executor.south) -- (evaluation.north); \draw[flow] (evaluation) -- (refiner);
\draw[flow,dashed] (refiner.west) -- ++(-0.45,0) |- (executor.south);

\end{tikzpicture}
\caption{Conceptual synthesis of adaptive educational functions and the prompt-refinement loop. Content preparation and course-grounding functions were informed by prior technical-learning systems \cite{raul2025ragprism,patel2025ocr}, while dialogue analyses informed learner-state representations \cite{lin2026llmmcaffect}. The present experiments evaluated the dashed model-level refinement loop on affective and cognitive tasks. They did not evaluate integration with a course platform, a deployed retrieval service, or student learning outcomes.}
\label{fig:adaptive_arch}
\end{figure*}
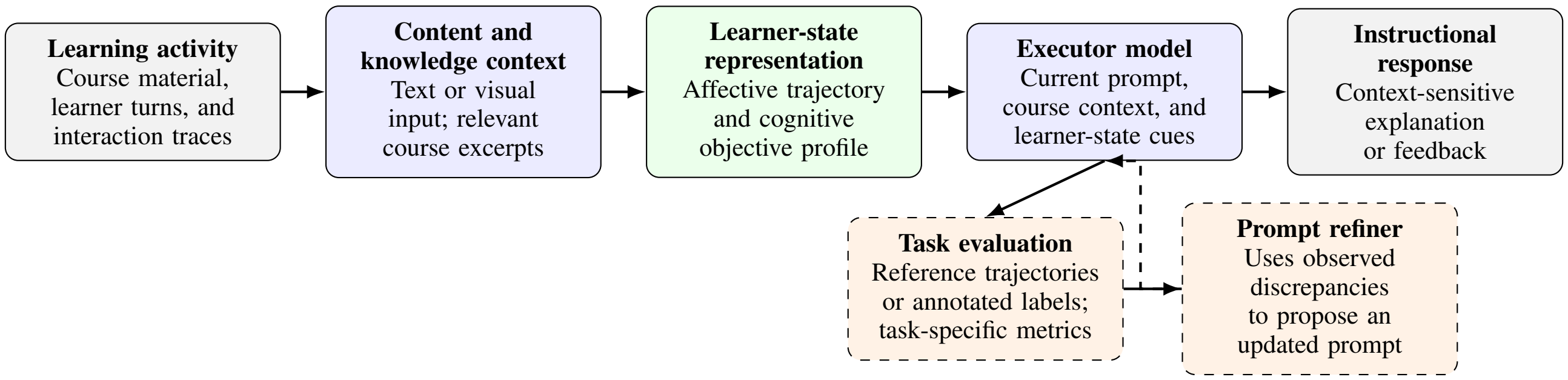

Figure~\ref{fig:adaptive_arch} synthesizes instructional-content preparation, course-knowledge access, learner-state representation, and context-sensitive response generation. The experiments have evaluated only the model-level prompt-refinement loop, represented by dashed feedback components; content preparation, knowledge access, and platform delivery have supplied architectural context without being integrated or experimentally evaluated here.

The interconnected demands of Industry 4.0 workforce education have motivated the selection of two complementary analytical tasks within this architecture: affective trajectory estimation, which represents sentiment as an instructional dialogue unfolds, and multi-label cognitive classification, which characterizes the demands articulated in learning outcomes. For equipment operation, fault diagnosis, and system-level reasoning, these representations could inform how instructional support responds to expressed difficulty and how learning objectives combine multiple cognitive processes. The experimental design has therefore examined whether a shared prompt-refinement procedure can improve these underlying analyses through distinct feedback signals, establishing a basis for subsequent investigation of their contribution to adaptive technical instruction.

PRISM has provided a conceptual precedent for combining learner dialogue and digital-twin observations with sentiment analysis and knowledge-augmented guidance \cite{lin2026prism}. Photogrammetry-derived equipment models could supply visual context \cite{alhamadah2024photogrammetry}, while generative digital-twin methods have represented evolving process behavior from structured observations \cite{lin2026dddgendt}. A proposed semiconductor-training extension of PRISM describes equipment-specific exercises in fabrication procedures, operation, and troubleshooting \cite{encinas2026semiconductor}. These examples identify potential deployment contexts. Their integration with the refinement loop and their educational effectiveness are outside the scope of this evaluation.

\par Algorithm~\ref{alg:rag_prism}, adapted from prior work \cite{raul2025ragprism}, specifies how learner context, retrieved course material, and response evaluation could connect within the architecture. Its retrieval and tutoring stages were not executed in the present experiments.

\begin{algorithm}[!t]
\caption{Retrieval-grounded adaptive tutoring (adapted from \cite{raul2025ragprism})}
\label{alg:rag_prism}
\begin{algorithmic}[1]
\REQUIRE Course documents $\mathcal{D}$; learner interaction $u_t$ and history $H_t$; sentiment estimator $g$; retriever $r$; language model $f$; synthetic query set $Q_s$; manually curated query set $Q_m$.
\ENSURE Grounded instructional response $a_t$ and retrieval/generation evaluation results.
\STATE Parse $\mathcal{D}$ into passages and construct a searchable vector index using document embeddings.
\STATE $s_t \leftarrow g(H_t,u_t)$ \COMMENT{Estimate learner affect/context from the interaction}
\STATE $q_t \leftarrow \textsc{FormQuery}(u_t,H_t,s_t)$
\STATE $C_t \leftarrow \textsc{RetrieveTopK}(q_t,r(\mathcal{D}))$
\STATE $a_t \leftarrow f(\textsc{Compose}(q_t,s_t,C_t))$ \COMMENT{Generate a response grounded in retrieved course passages}
\FORALL{$Q \in \{Q_s,Q_m\}$}
 \STATE Retrieve passages for each query in $Q$ and compare rankings with its reference passage
 \STATE Compute document Hit Rate and Mean Reciprocal Rank (MRR)
 \STATE Generate answers and assess source faithfulness and query relevancy
\ENDFOR
\RETURN $a_t$ and the retrieval/generation evaluation results
\end{algorithmic}
\end{algorithm}

\subsection{Problem Formulation and Task Definitions} Let $x$ denote an educational input, $P_t$ the executor prompt at refinement iteration $t$, and $f(P_t,x;\omega)$ the executor response under decoding randomness $\omega$. The study sought prompts that improved task-specific empirical performance on refinement data, without assuming an explicit global objective or a guarantee that successive revisions converged. In the affective task, a dialogue comprised an ordered sequence of utterances $c=(u_1,\ldots,u_n)$, and the executor assigned each utterance a score $s_i\in[0,5]$, where lower values denoted more positive sentiment, higher values denoted more negative sentiment, and $2.5$ represented a neutral baseline. For trajectory analysis, the scores were mapped to $[-1,1]$ according to $z_i=1-2(s_i/5)$ and collected as $z=(z_1,\ldots,z_n)$. A deterministic LLM-generated trajectory $y^{\mathrm{ref}}$ served as the refinement and evaluation reference, providing a reproducible target but not independently validated human affect labels. For cognitive classification, each learning outcome $x$ was paired with a six-dimensional binary label vector $y\in\{0,1\}^{6}$, whose entries represented the annotated presence or absence of the six Bloom's Taxonomy levels; the executor generated a corresponding prediction $\hat{y}$.\subsection{Closed-Loop Iterative Refinement} At each iteration, the executor processed refinement inputs using the current prompt, after which the refiner received the prompt and a structured description of the observed discrepancy and proposed a revised prompt. The update was represented as $P_{t+1}=g(I_t,P_t)$, where $g$ denoted the refiner and $I_t$ the task-specific feedback; no executor weights were updated. For the affective task, feedback comprised the dialogue, the mean trajectory estimated from repeated executor generations, and the deterministic reference trajectory, $I_t^{\mathrm{aff}}=(c,\bar{y}_t,y^{\mathrm{ref}})$, with $\bar{y}_t=K^{-1}\sum_{k=1}^{K} f(P_t,c;\omega_k)$, where $K$ denotes the number of independent executor generations used to estimate the mean trajectory. The refiner used differences between $\bar{y}_t$ and $y^{\mathrm{ref}}$ to revise the instruction, thereby targeting reference alignment while repeated generations exposed variation attributable to stochastic decoding. For cognitive classification, feedback comprised the refinement examples on which predictions disagreed with labels, $I_t^{\mathrm{cog}}=E_t=\{(x_i,\hat{y}_{i,t},y_i):\hat{y}_{i,t}\ne y_i\}$; the refiner used these error instances to clarify the prompt's classification criteria and output requirements. Algorithms \ref{alg:affective_refinement} and \ref{alg:bloom_refinement} specified the two task-specific feedback loops evaluated in this study. Here, $T$ and $K$ denote the configured upper bound on refinement rounds and the number of executor generations used to estimate each mean affective trajectory, respectively. We selected the best candidate using the task-specific refinement-set metric, mean absolute error for trajectory alignment and micro-F1 for multi-label classification, while held-out examples remained outside prompt revision and selection. Retaining the best observed candidate provided a selection rule, but did not constitute evidence of convergence or monotonic improvement across updates.

\begin{algorithm}[!t]
\caption{Affective trajectory prompt refinement}
\label{alg:affective_refinement}
\begin{algorithmic}[1]
\REQUIRE Initial prompt $P_0$, refinement dialogues $\mathcal{C}$, executor $f$, refiner $g$, deterministic reference model $r$, repeated-run count $K$, iteration bound $T$.
\ENSURE Best prompt $P^*$ observed during refinement.
\STATE $P^* \leftarrow P_0$; $m^* \leftarrow +\infty$
\FORALL{$c \in \mathcal{C}$}
 \STATE $y^{\mathrm{ref}}(c) \leftarrow r(c)$
\ENDFOR
\FOR{$t=0$ to $T-1$}
 \FORALL{$c \in \mathcal{C}$}
  \STATE Generate $K$ trajectories with $f(P_t,c)$ and compute their mean $\bar{y}_t(c)$
  \STATE Form feedback $I_t^{\mathrm{aff}}(c)=(c,\bar{y}_t(c),y^{\mathrm{ref}}(c))$
 \ENDFOR
 \STATE Compute $m_t$, the refinement-set MAE between $\bar{y}_t(c)$ and $y^{\mathrm{ref}}(c)$ over $\mathcal{C}$
 \IF{$m_t<m^*$}
  \STATE $P^* \leftarrow P_t$; $m^* \leftarrow m_t$
 \ENDIF
 \STATE $P_{t+1} \leftarrow g(\{I_t^{\mathrm{aff}}(c):c\in\mathcal{C}\},P_t)$
\ENDFOR
\RETURN $P^*$
\end{algorithmic}
\end{algorithm}

\begin{algorithm}[!t]
\caption{Multi-label Bloom taxonomy prompt refinement}
\label{alg:bloom_refinement}
\begin{algorithmic}[1]
\REQUIRE Initial prompt $P_0$, labeled refinement data $\mathcal{D}$ processed in partitions, executor $f$, refiner $g$, iteration bound $T$.
\ENSURE Best prompt $P^*$ observed during refinement.
\STATE $P^* \leftarrow P_0$; $F^* \leftarrow -\infty$
\FOR{$t=0$ to $T-1$}
 \STATE $E_t \leftarrow \emptyset$; collect predictions and labels across refinement partitions
 \STATE Generate multi-label predictions $\hat{y}_{i,t}$ with $f(P_t,x_i)$ for each $(x_i,y_i)\in\mathcal{D}$
 \STATE Compute aggregate micro-F1 $F_t$ and collect errors $E_t=\{(x_i,\hat{y}_{i,t},y_i):\hat{y}_{i,t}\ne y_i\}$
 \IF{$F_t>F^*$}
  \STATE $P^* \leftarrow P_t$; $F^* \leftarrow F_t$
 \ENDIF
 \STATE $P_{t+1} \leftarrow g(E_t,P_t)$
\ENDFOR
\RETURN $P^*$
\end{algorithmic}
\end{algorithm}

\begin{figure*}[!t]
\centering
\includegraphics[width=0.90\textwidth]{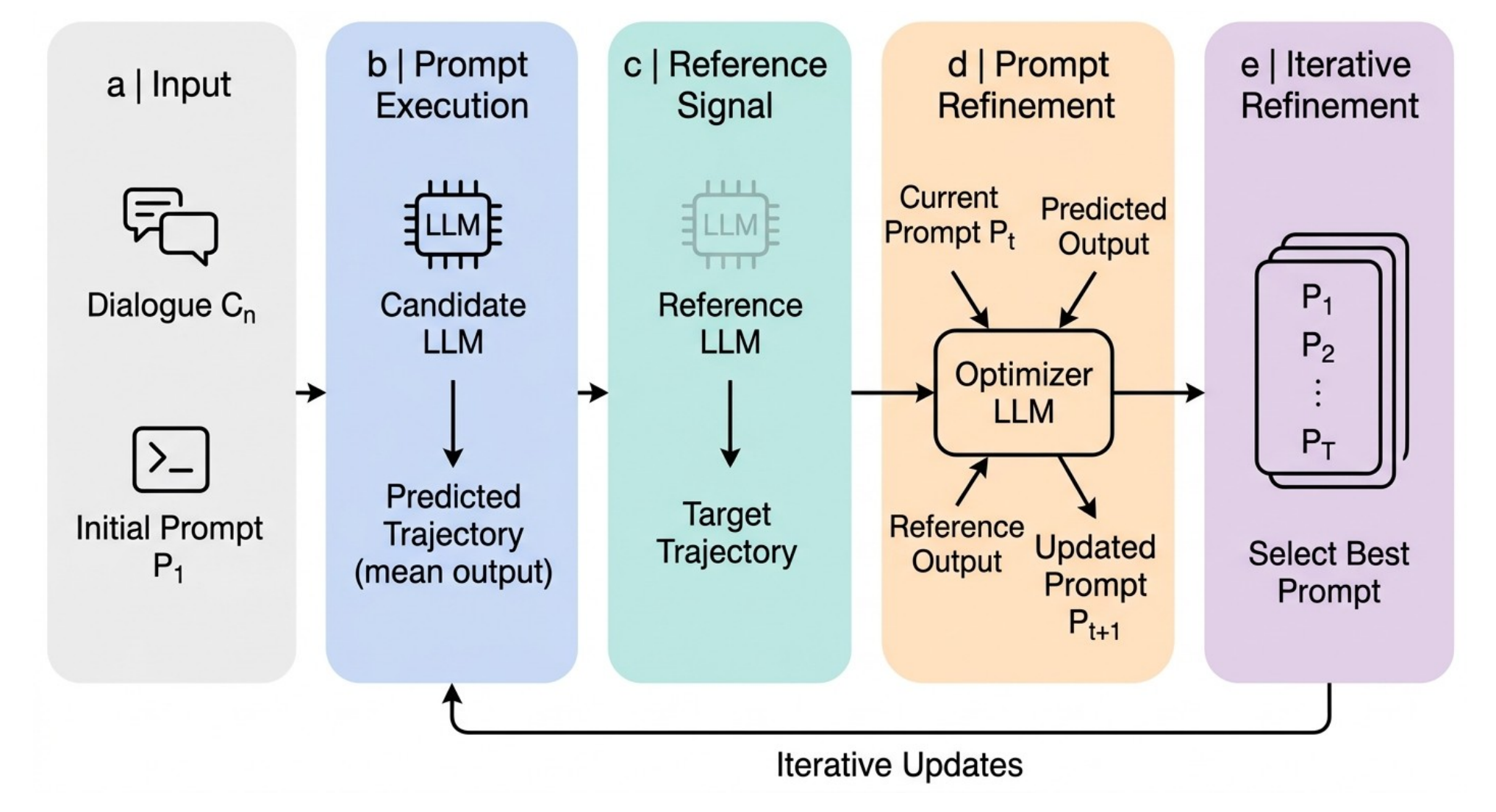}
\caption{Prompt-refinement workflow evaluated in this study. The affective procedure compared executor trajectories with a deterministic reference, and the cognitive procedure used discrepant predictions to guide prompt revision.}
\label{fig:refinement_workflow}
\end{figure*}

\begin{figure*}[!t]
\centering
\includegraphics[width=0.82\textwidth]{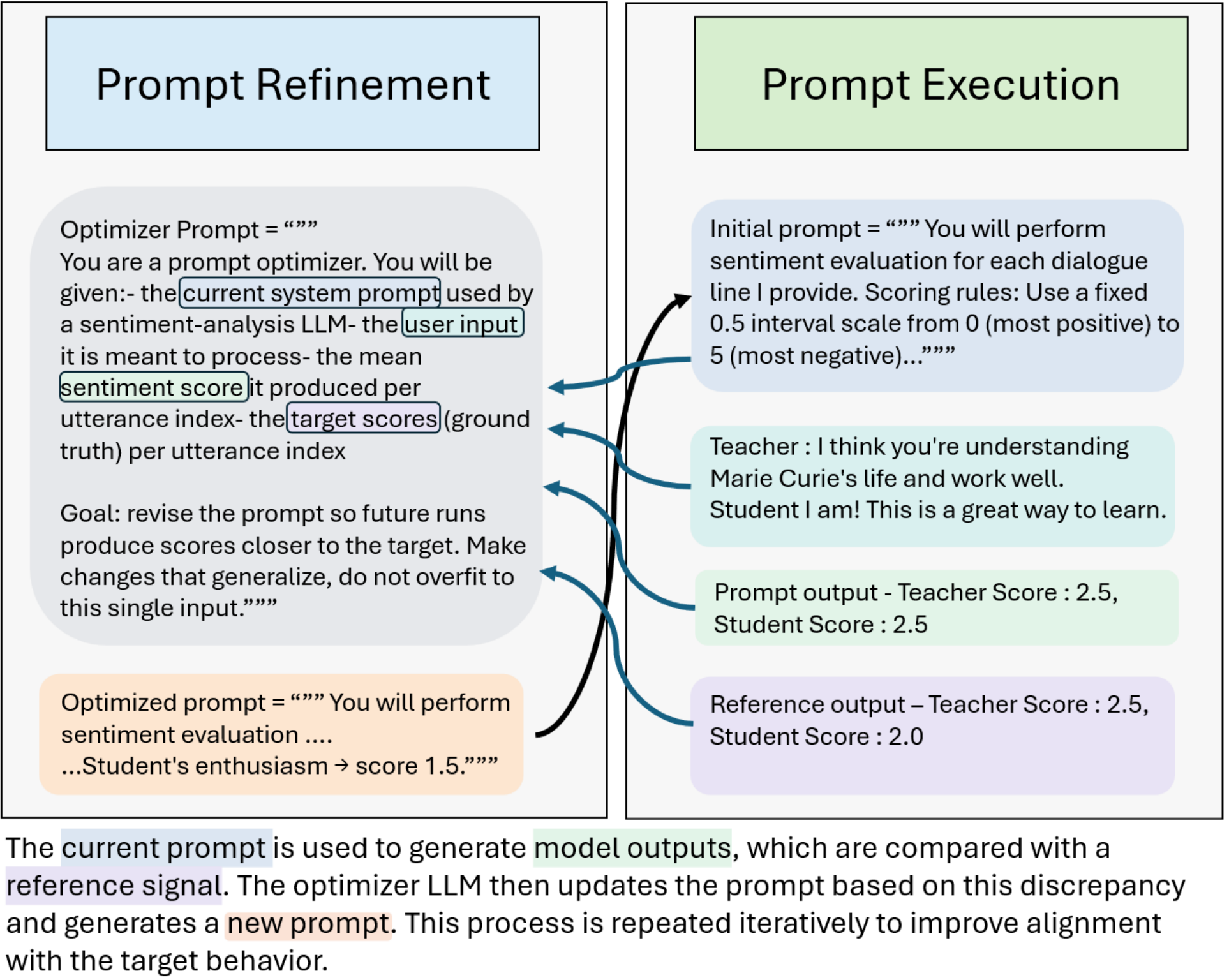}
\caption{Example of a single affective prompt-refinement iteration. The executor's response is compared with a deterministic reference, and the observed discrepancy informs the refiner's update to the executor prompt. The corresponding cognitive-task feedback uses discrepant labeled examples, as specified in Algorithm~\ref{alg:bloom_refinement}.}
\label{fig:single_iteration_refinement}
\end{figure*}

Figure~\ref{fig:single_iteration_refinement} illustrates affective feedback within one refinement cycle, whereas Algorithm~\ref{alg:bloom_refinement} specifies the distinct labeled-error feedback used for cognitive classification. Neither representation establishes convergence across iterations.

\subsection{Refiner-System Instructions}
\label{sec:refiner_prompts}
The following meta-prompts guided the refiner's revisions of executor instructions. The Bloom procedure supplied discrepant labeled examples, whereas the affective procedure supplied dialogue, mean predicted trajectory, and reference trajectory; these meta-prompts were distinct from the executor prompts being revised.

\subsection*{Bloom's Taxonomy classification refiner}\begin{quote}You are a prompt optimizer for a Bloom's Taxonomy multi-label classifier.\par\medskip You will be given: the current system prompt used by the classifier LLM; and learning outcomes for which the classifier's majority-vote prediction does not match ground truth, with both the predicted and ground-truth binary label vectors shown.\par\medskip Goal: revise the prompt to improve multi-label Bloom's Taxonomy classification. Encourage the model to consider all applicable levels and avoid over- or under-prediction. Make changes that generalize; do not overfit to the specific examples shown.\par\medskip Rules: output only the revised system prompt, with no explanation, preamble, or commentary; preserve the output JSON format exactly, with an array of results containing \texttt{id} and \texttt{labels} fields, and ensure that \texttt{labels} lists all applicable Bloom's levels; content may be restructured, added, or removed.\end{quote}

\subsection*{Sentiment-trajectory refiner}\begin{quote}You are a prompt optimizer. You will be given the current system prompt used by a sentiment-analysis LLM, the user input it is meant to process, the mean sentiment score it produced per utterance index, and the target scores (ground truth) per utterance index.\par\medskip Goal: revise the prompt so future runs produce scores closer to the target. Make changes that generalize; do not overfit to this single input.\par\medskip Rules: output only the revised system prompt, with no explanation or preamble; you may restructure, add, or remove content; preserve the JSON format instructions exactly.\end{quote}

\subsection{Data and Experimental Design}
The affective task used EduTalk, approximately 1,280 teacher--student dialogues annotated for conversation-level binary sentiment \cite{lin2025personalized}; the present trajectory task instead used utterance-level model-generated references. The cognitive task used multi-label learning outcomes from the dataset reported by Li et al. \cite{li2022bloom}. Table~\ref{tab:task_configuration} specifies the models, temperatures, refinement and evaluation sample sizes, feedback, and measures. Both tasks were evaluated zero-shot without task-specific fine-tuning, and held-out examples supplied no refinement feedback. Four affective refinement conversations did not overlap with the 100 test conversations, while 1,000 Bloom outcomes were reserved separately from the 70 refinement examples. The records do not specify sampling procedures, Bloom refinement-partition sizes, label frequencies, or split-generation methods; consequently, interpretation remains limited to the reported samples. Section~\ref{sec:refiner_prompts} preserves the refiner instructions, and the table note identifies remaining replication limits.

\begin{table*}[!t]
\centering
\caption{Experimental configurations for the affective and cognitive prompt-refinement tasks.}
\label{tab:task_configuration}
\small
\setlength{\tabcolsep}{4pt}
\begin{tabular}{|p{0.13\textwidth}|p{0.40\textwidth}|p{0.39\textwidth}|}
\hline
\textbf{Task} & \textbf{Models and refinement feedback} & \textbf{Data and evaluation} \\
\hline
Affective trajectory alignment & Phi-4 executor and Trinity-Large-Preview refiner (temperature 0.7); Nemotron-3-Nano-30B-A3B deterministic reference (temperature 0.0). Feedback comprised each dialogue, the mean trajectory over repeated executor generations, and the reference trajectory. & Four conversations informed refinement; a separate, non-overlapping set of 100 conversations supported held-out evaluation. Outcomes included mean absolute error against the reference and mean output standard deviation across repeated generations. \\
\hline
Bloom's Taxonomy multi-label classification & GPT-OSS-120B executor and Trinity-Large-Preview refiner (temperature 0.7). Feedback comprised refinement examples for which the predicted label vector differed from the annotated vector. & Seventy labeled examples were processed in sequential refinement partitions; 1,000 separate outcomes supported held-out evaluation. Outcomes included micro-F1 and exact-match subset accuracy across six labels. \\
\hline
\end{tabular}
\vspace{2pt}
\parbox{0.94\textwidth}{\footnotesize Both tasks were evaluated zero-shot without task-specific fine-tuning, and held-out examples did not supply refiner feedback. The full held-out evaluation did not report its repeated-generation count, and it did not specify the refinement-round bound. We denote these quantities by $K$ and $T$. The case figures show ten inference attempts for two selected dialogues, but the record does not establish that this count was used for the full evaluation. Model-service revisions and access dates, random seeds, decoding settings beyond temperature, retry rules, a detailed prompt-selection tie-break, and exact initial and selected executor prompts were also unavailable, limiting exact replication. Section~\ref{sec:refiner_prompts} provides the refiner meta-prompts.}
\end{table*}

\subsection{Evaluation Measures} For the affective task, the mean executor prediction across repeated generations supplied the trajectory used to calculate mean absolute error (MAE) against the deterministic LLM-generated reference; average standard deviation across held-out conversations summarized generation variability. For cognitive classification, we averaged predictions across repeated executor runs. Micro-F1 across the six Bloom's Taxonomy labels served as the primary measure, subset accuracy required an exact match across all labels for an outcome, and standard deviation summarized run-to-run variability. The reported measures therefore captured reference agreement, multi-label classification performance, and output variability; the affective reference did not constitute independent human ground truth, and neither task measured instructional efficacy, retention, or achievement.

\section{Experimental Results and Analysis}
The evaluation has examined two analytical functions relevant to adaptive Industry 4.0 workforce education. It represents sentiment across instructional dialogue and characterizes the cognitive demands of learning outcomes. Comparisons between initial and selected prompts have assessed whether feedback-driven refinement improves reference agreement, classification performance, and generation consistency on held-out inputs, providing evidence about the model behavior underlying these functions. Although equipment operation, fault diagnosis, and system-level reasoning have motivated their potential application, the evaluated dialogues and learning outcomes have supported task-level analysis rather than direct assessment of workforce training.

\subsection{Sentiment Trajectory Alignment Results} On a held-out test split of 100 non-overlapping student dialogues from the EduTalk Sentiment Dataset, the optimized prompt $P_{opt}$ reduced the Mean Absolute Error (MAE) relative to the deterministic reference trajectory from 0.742 to 0.702, representing a 5.39\% relative error reduction. Simultaneously, the average standard deviation ($\sigma$) across independent stochastic decoding runs decreased from 0.307 to 0.238, representing a 22.48\% reduction in output variability. Table~\ref{tab:sentiment_results} summarizes these quantitative metrics. \begin{table}[!t]
\centering
\caption{Sentiment Trajectory Alignment Performance on the Held-Out Set ($N=100$ Conversations)}
\label{tab:sentiment_results}
\small
\begin{tabular}{lcc}
\hline
Metric & Baseline & Optimized \\
\hline
MAE & 0.742 & \textbf{0.702} \\
Mean generation SD ($\sigma$) & 0.307 & \textbf{0.238} \\
\hline
\end{tabular}
\end{table} Lower MAE indicated closer agreement with the deterministic Nemotron reference, rather than independently validated affective accuracy. Lower generation standard deviation indicated greater consistency, not uncertainty about a learner's true affect. These aggregate measures did not establish reliable detection of particular emotional transitions or improved instructional decisions. 

\subsection{Case-Based Affective-Trajectory Analysis}
Two selected EduTalk dialogues provided case-based evidence of the trajectory patterns and run-to-run variation summarized by the aggregate results. In the Cold War exchange, the mean sentiment estimates for the student and teacher declined together; in the French Revolution exchange, the student trajectory shifted from negative to positive during the narrative-oriented interaction. Shaded regions show standard deviations across ten repeated sentiment-analysis attempts for each dialogue. These repeated generations characterize output variability and are distinct from prompt-refinement iterations. The cases provide descriptive context for the affective task. However, they do not add independent observations to the held-out sample or establish that the instructional strategy caused the observed changes or improved learning. These history-based dialogues have illustrated the affective-analysis procedure in general educational interactions. Their relevance to Industry 4.0 instruction rests on the proposed analytical function, while their validity for equipment-specific dialogue and technical troubleshooting remains to be established.

\begin{figure*}[!t]
\centering
\includegraphics[width=0.92\textwidth,height=0.72\textheight,keepaspectratio]{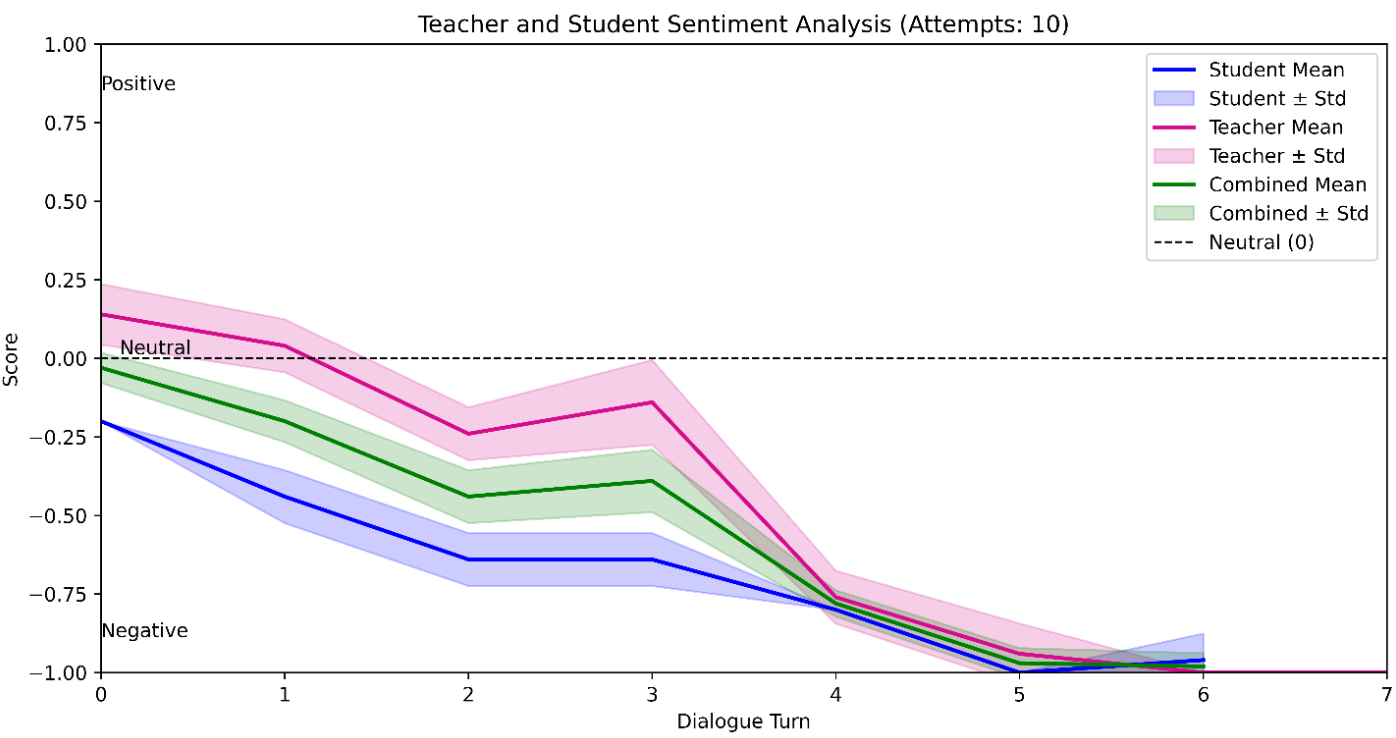}
\caption{Estimated sentiment trajectories across successive utterances in a Cold War instructional dialogue. Curves show student, teacher, and combined mean scores; shaded bands indicate standard deviations across ten repeated inference attempts. These are model-generated sentiment estimates, with positive values indicating more positive sentiment.}
\label{fig:coldwar_sentiment}
\end{figure*}

\begin{figure*}[!t]
\centering
\includegraphics[width=0.92\textwidth,height=0.72\textheight,keepaspectratio]{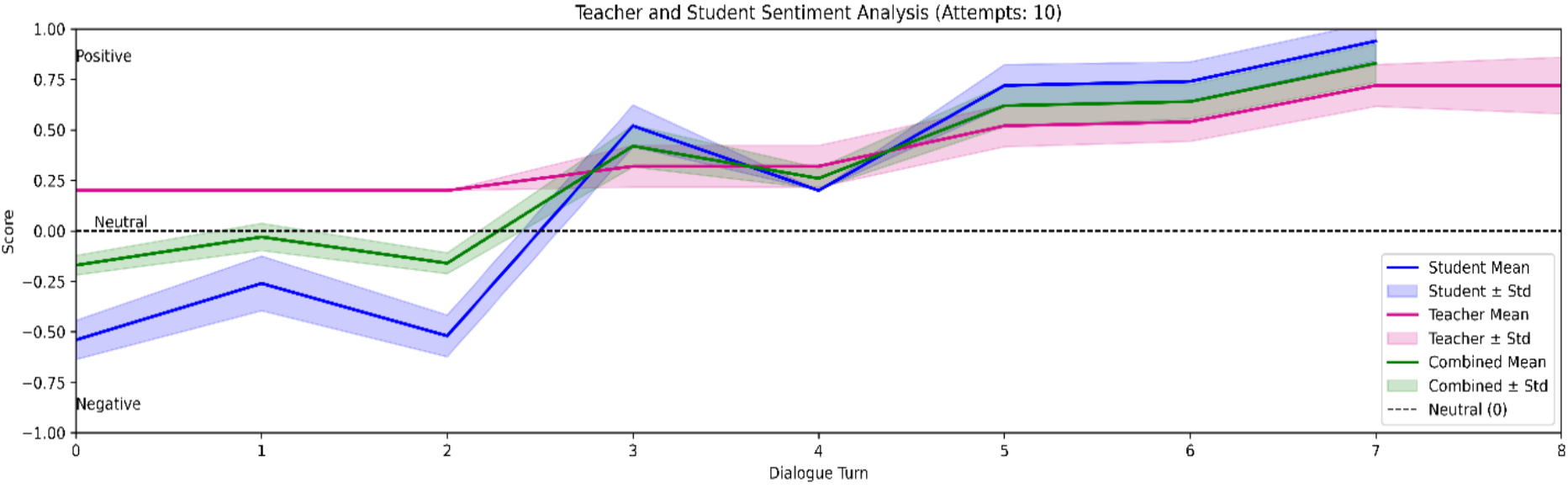}
\caption{Estimated sentiment trajectories across successive utterances in a French Revolution instructional dialogue. The student's mean trajectory shifts from negative to positive during the exchange; shaded bands indicate standard deviations across ten repeated inference attempts. The observed pattern does not establish an instructional effect.}
\label{fig:french_sentiment}
\end{figure*}

\subsection{Bloom's Taxonomy Multi-Label Classification Results} For the multi-label cognitive taxonomy classification task evaluated on a held-out test split of 1,000 learning outcome statements, the optimized prompt $P_{opt}$ achieved a micro-F1 score of 0.893 compared with 0.687 for the baseline prompt $P_0$, representing an absolute improvement of 0.206 (a 30.00\% relative increase). Concurrently, subset exact-match accuracy increased from 0.418 to 0.636, representing an absolute improvement of 0.218 (a 52.15\% relative increase). Table~\ref{tab:blooms_results} presents these classification metrics. \begin{table}[!t]
\centering
\caption{Multi-Label Bloom Classification Performance on the Held-Out Set ($N=1{,}000$ Outcomes)}
\label{tab:blooms_results}
\small
\begin{tabular}{lcc}
\hline
Metric & Baseline & Optimized \\
\hline
Micro-F1 & 0.687 & \textbf{0.893} \\
Subset accuracy & 0.418 & \textbf{0.636} \\
\hline
\end{tabular}
\end{table} The increases in micro-F1 and subset accuracy were consistent with improved multi-label performance on the held-out outcomes. These aggregate measures, however, did not identify which Bloom levels or error types accounted for the change, and the present analysis did not isolate the contribution of individual prompt revisions. Accordingly, the results supported an association between the selected prompt and higher held-out scores under this evaluation protocol, rather than a causal claim about particular criteria or a general claim of error mitigation across technical learning tasks. 

\begin{figure*}[!t]
\centering
\includegraphics[width=0.88\textwidth]{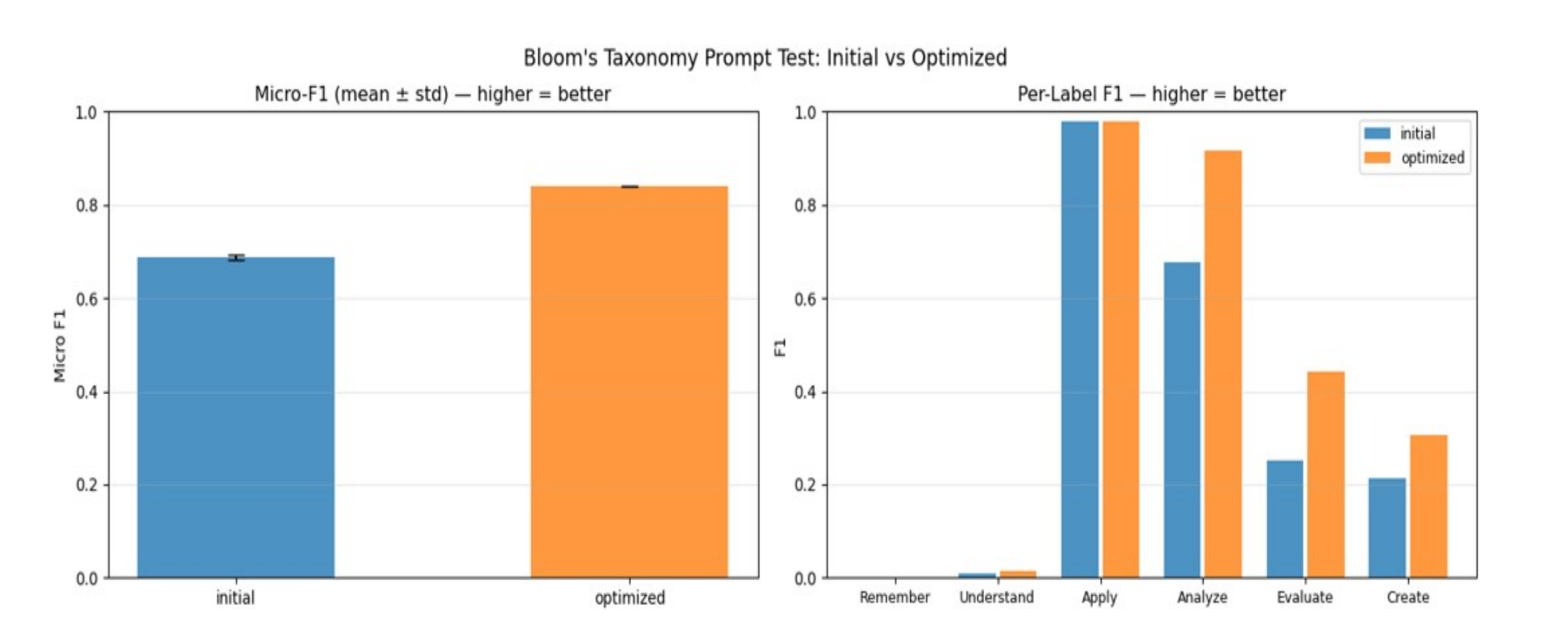}
\caption{Initial and optimized prompt performance for multi-label Bloom classification on the held-out set of 1,000 learning outcomes. The left panel reports micro-F1; the right panel displays per-label F1 values reported for the evaluated outcomes. The figure supports descriptive comparison and does not provide additional independently collected observations.}
\label{fig:bloom_performance}
\end{figure*}

\subsection{Iteration Evidence and Interpretation}
The results have established endpoint comparisons rather than a convergence trajectory: no per-iteration performance series was available to identify monotonic improvement, convergence timing, or stable plateaus. Figure~\ref{fig:bloom_performance} compares initial and selected Bloom prompts, while Figs.~\ref{fig:coldwar_sentiment} and~\ref{fig:french_sentiment} characterize inference variability for selected dialogues. Their ten repeated generations do not represent ten prompt-refinement rounds.

\section{Discussion}
The held-out comparisons have shown closer agreement with a model-generated affective reference, lower affective-generation variability, and higher multi-label Bloom classification scores following prompt refinement without executor parameter updates. These findings have supported endpoint improvements across two educational analyses, while leaving human-valid affect estimation and educational effectiveness unresolved.

\subsection{Interpretation in Relation to Prompt Optimization Literature}
The endpoint improvements have been consistent with the model-generated search and feedback-guided revision mechanisms of APE, OPRO, and Self-Refine \cite{zhou2022large,yang2023large,madaan2023self}. However, the study has not compared optimizers, expert-authored prompts, or fine-tuning, or isolated individual prompt changes; relative superiority and the causes of improvement therefore remain unresolved. Course-grounded immersive tutoring \cite{raul2025ragprism} has addressed access to domain evidence, whereas the present loop has targeted interpretation of task instructions and feedback. Their integration remains a hypothesis for future evaluation.

\subsection{Cost and Measurement Scope}
Refinement has required additional executor and refiner calls, but timing and token-cost measurements were insufficient to quantify overhead or assess real-time suitability. Findings from a separate slide-processing pipeline do not quantify refinement costs \cite{patel2025ocr}. Future evaluations should report end-to-end latency, token use, and prompt-update frequency alongside task performance.

\subsection{Validity of Model-Generated and Annotated Targets} The affective task used a deterministic Nemotron reference, making the comparison reproducible. This use of a fixed reference differs from LLM-MC-Affect's use of repeated inferences to characterize trajectory variation \cite{lin2026llmmcaffect}. Neither approach provides objective or human-validated ground truth. The measured reduction in error therefore indicated closer agreement with that reference, not improved diagnosis of learners' emotional states. Reduced generation variability indicates greater output consistency but does not establish improved affective validity. Because repeated-inference dispersion may reflect both model stochasticity and meaningful interpretive ambiguity, future evaluation should test whether prompt refinement preserves disagreement observed among human raters \cite{lin2026llmmcaffect}. The Bloom task used labels associated with Li et al.'s learning-outcome dataset. Li et al. reported coding by a trained annotator and independent review of 30\% of the objectives, with Cohen's $\kappa$ increasing from 0.63 to 0.80 after label revision \cite{li2022bloom}. These source-level checks support annotation provenance but do not independently establish the validity or label distribution of the subset evaluated here. Follow-up studies should use expert review, multiple human raters, inter-rater agreement, and subgroup error analyses before using either signal to support consequential instructional decisions. 

Exploratory research on agent-generated textual artifacts has combined model-assisted interpretation with subsequent human verification while acknowledging influences from prompts, model providers, and human intervention \cite{lin2026siliconsocieties}. These methodological limits reinforce the need to independently validate educational interpretations before using model-derived signals to guide instruction.

\subsection{Implications for Adaptive Education} Affective signals and adaptive prompts may eventually support responsive instruction, but the present experiments measured model outputs rather than learner engagement, persistence, retention, or equitable outcomes. An earlier PPE-training pilot has reported higher task hit rates and shorter completion times with adaptive difficulty, although unequal participant groups and repeated observations have limited causal interpretation \cite{lin2025personalized}. Prior work situated personalized tutoring and technical instruction within educational contexts \cite{raul2025ragprism,lin2026llmmcaffect,patel2025ocr}, but neither those studies nor the present offline evaluation established that sentiment-based adaptation improves learning. Such use would require learner-centered validation, transparent instructor oversight, and assessment of differential errors before affective inference informed instructional decisions.

\subsection{Limitations and Future Research}
The evidence has remained limited to endpoint comparisons on fixed held-out sets and small refinement samples, without independent repetitions of the full optimization procedure, random-seed analyses, confidence intervals, or inferential tests. Generation standard deviation has not supplied a confidence interval for the measured improvement, and neither human validation of affect estimates nor learner outcomes has been assessed. Follow-up studies should log each prompt update, model version, prompt, sampling decision, decoding setting, and call budget, and report independently repeated optimization runs with confidence intervals. Comparisons with expert-authored prompts and established optimizers should use matched token and latency budgets. Human-rated affect with agreement statistics, audited Bloom annotations and per-class errors, and supervised laboratory pilots measuring task completion and learner feedback should precede adequately powered multi-site studies of retention and transfer.

\section{Conclusion}
Feedback-driven prompt refinement has improved measured performance without executor parameter updates: across 100 EduTalk conversations, reference-based sentiment MAE has decreased from 0.742 to 0.702 and generation standard deviation from 0.307 to 0.238; across 1,000 learning outcomes, Bloom micro-F1 has increased from 0.687 to 0.893 and subset accuracy from 0.418 to 0.636. These task-level findings have complemented course-grounded educational architectures \cite{raul2025ragprism} without establishing human-valid affect recognition, optimization convergence, or learning gains. Human validation, optimizer comparisons, and supervised learner studies remain necessary to determine whether these analytical improvements support adaptive Industry 4.0 workforce education.

\bibliographystyle{IEEEtran}
\bibliography{Bibliography}

\end{document}